\documentclass{article}
\usepackage{ijcai23}

\usepackage{times}
\usepackage{soul}
\usepackage{url}
\usepackage[hidelinks]{hyperref}
\usepackage[utf8]{inputenc}
\usepackage[small]{caption}
\usepackage{graphicx}
\usepackage{amsmath}
\usepackage{amsthm}
\usepackage{booktabs}
\usepackage{algorithm}
\usepackage{algorithmic}
\usepackage[switch]{lineno}

\usepackage{etoolbox}  
\makeatletter  
\patchcmd{\thebibliography}{\chapter*}{\section*}{}{}  
\makeatother  

\title{Video Diffusion Models with Local-Global Context Guidance}

\author{
Siyuan Yang $^1$
\and
Lu Zhang$^2$ \and
Yu Liu$^1$\thanks{Corresponding authors} \and
Zhizhuo Jiang $^1$ \And
You He $^1$
\affiliations
$^1$Tsinghua University \\
$^2$Dalian University of Technology\\
\emails
yang-sy21@mails.tsinghua.edu.cn,
zhangluu@dlut.edu.cn,
\{liuyu77360132, heyou\_f\}@126.com,
jiangzhizhuo@sz.tsinghua.edu.cn
}

\begin{document}

\maketitle

\begin{abstract}
Diffusion models have emerged as a powerful paradigm in video synthesis tasks including prediction, generation, and interpolation. Due to the limitation of the computational budget, existing methods usually implement conditional diffusion models with an autoregressive inference pipeline, in which the future fragment is predicted based on the distribution of adjacent past frames. However, only the conditions from a few previous frames can't capture the global temporal coherence, leading to inconsistent or even outrageous results in long-term video prediction.  
In this paper, we propose a Local-Global Context guided Video Diffusion model (LGC-VD) to capture multi-perception conditions for producing high-quality videos in both conditional/unconditional settings. In LGC-VD, the UNet is implemented with stacked residual blocks with self-attention units, avoiding the undesirable computational cost in 3D Conv. We construct a local-global context guidance strategy to capture the multi-perceptual embedding of the past fragment to boost the consistency of future prediction. Furthermore, we propose a two-stage training strategy to alleviate the effect of noisy frames for more stable predictions. Our experiments demonstrate that the proposed method achieves favorable performance on video prediction, interpolation, and unconditional video generation. We release code at \url{https://github.com/exisas/LGC-VD}.
\end{abstract}

\section{Introduction}
\begin{figure}[t]
\centering
\includegraphics[width=\linewidth]{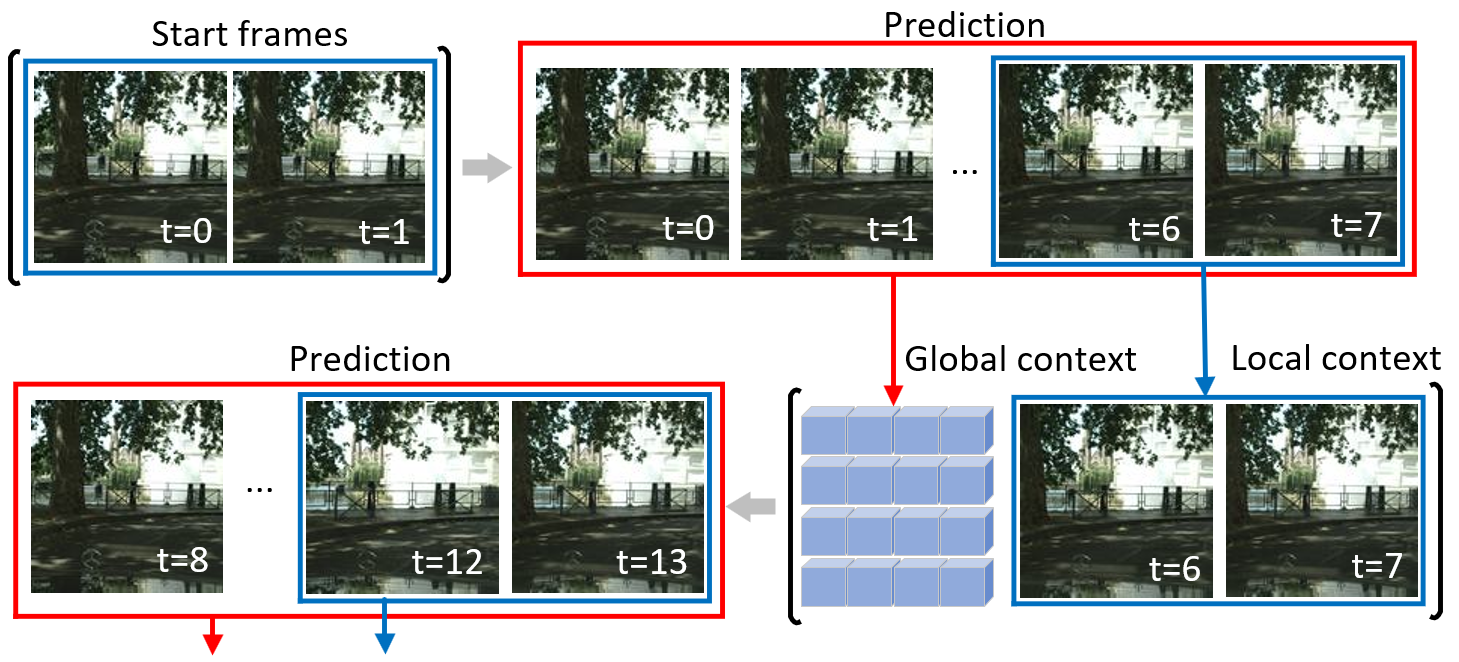}
\caption{
We propose an autoregressive inference framework, where both global context and local context from previous predictions are incorporated to enhance the consistency of the next video clip.
}
\label{condition}
\end{figure}
Video prediction aims to generate a set of future frames that are visually consistent with the past content. By capturing the future perceive in a dynamic scene, video prediction has shown great value in many applications such as automaton driving~\cite{hu2020probabilistic} and human-machine interaction~\cite{wu2021greedy}. To make the generated videos more aesthetic, most of the early deep learning methods rely on 3D convolution~\cite{tran2015learning} or RNNs~\cite{vRNN1,vRNN2} to model the temporal coherence of historical frames. However, due to limited long-term memory capacity in both 3D Conv and RNNs, these methods fail to capture the global correspondence of past frames, leading to the inconsistency of both semantics and motion in future frame prediction.

Inspired by the success of Generative Adversarial Networks (GANs) in image synthesis~\cite{GAN1}, more recent attempts~\cite{GAN6,GAN8} have been made to develop video prediction frameworks on GANs by incorporating Spatio-temporal coherent designs. Thanks to the impressive generative ability of GANs, these methods show great effect in producing videos with natural and consistent content and hold the state-of-the-art in the literature. However, they still suffer from the inherent limitations of GANs in training stability and sample diversity~\cite{dhariwal2021diffusion}, which incur unrealistic texture and artifacts in the generated videos.  

Recently, Diffusion Probabilistic Models (DPMs) have gained increasing attention and emerged as a new state-of-the-art in video synthesis~\cite{voleti2022mcvd,ho2022video} and beyond~\cite{lugmayr2022repaint,rombach2022high}. Being likelihood-based models, diffusion models show strengths over GANs in training stability, scalability, and distribution diversity. The core insight is to transform the noise-to-image translation into a progressive denoising process, that is the predicted noise by a parameterized network should be approximate to the predefined distribution in the forward phase. For video prediction, an early attempt by \cite{ho2022video} is to extend the standard image diffusion model~\cite{ho2020denoising} with 3D UNet. The proposed Video Diffusion Model (VDM) thus can denoise the 3D input and return a cleaner video clip iteratively. Despite the impressive results, the sequential denoising mode and the 3D convolution are costly and largely limit the length and resolution of the predicted videos. To produce high-fidelity results, some following works~\cite{ho2022imagen,singer2022make} incorporate extra diffusion models for super-resolution, which further burden the computational cost during training and inference. 

To alleviate this issue, MCVD~\cite{voleti2022mcvd} uses stacked residual blocks with self-attention to form the UNet, so that the whole model can be trained and evaluated by limited computation resources within proper inference time. Besides, they propose an autoregressive inference framework where the masked context is taken as a condition to guide the generation of the next clip. As a result, MCVD can solve video prediction, interpolation, and unconditional generation with one unified framework. However, only a few adjacent frames (\emph{e.g.,} two past frames) are fed to the conditional diffusion model, without a global comprehension of the previous fragment. The model might be affected by the noisy past frames and produces inconsistent or even outrageous predictions in long-term videos. 

In this paper, we propose a Local-Global Context guided Video Diffusion model (LGC-VD) to capture comprehensive conditions for high-quality video synthesis. Our LGC-VD follows the conditional autoregressive framework of MCVD~\cite{voleti2022mcvd}, so the related tasks like video prediction, interpolation, and unconditional generation can be uniformly solved with one framework. As shown in Figure~\ref{condition}, we propose a local-global context guidance strategy, where both global context and local context from previous predictions are incorporated to enhance the consistency of the next video clip. Specifically, the local context from the past fragment is fed to the UNet for full interaction with the prediction during the iterative denoising process. A sequential encoder is used to capture the global embedding of the last fragment, which is integrated with the UNet via latent cross-attention. Furthermore, we build a two-stage training algorithm to alleviate the model's sensitivity to noisy predictions. 

Our main contributions are summarized as follows:

\begin{itemize}
  \item [1.] 
  A local-global context guidance strategy to capture the multi-perception conditions effectively for more consistent video generation.   
  \item [2.]
  A two-stage training algorithm, which treats prediction errors as a form of data augmentation, helps the model learn to combat prediction errors and significantly enhance the stability for long video prediction.
  \item [3.]
  Experimental results on two datasets demonstrate that the proposed method achieves promising results on video prediction, interpolation, and unconditional generation.  
\end{itemize}


\section{Related Work}
\begin{figure*}[h]
\centering
\includegraphics[width=\linewidth]{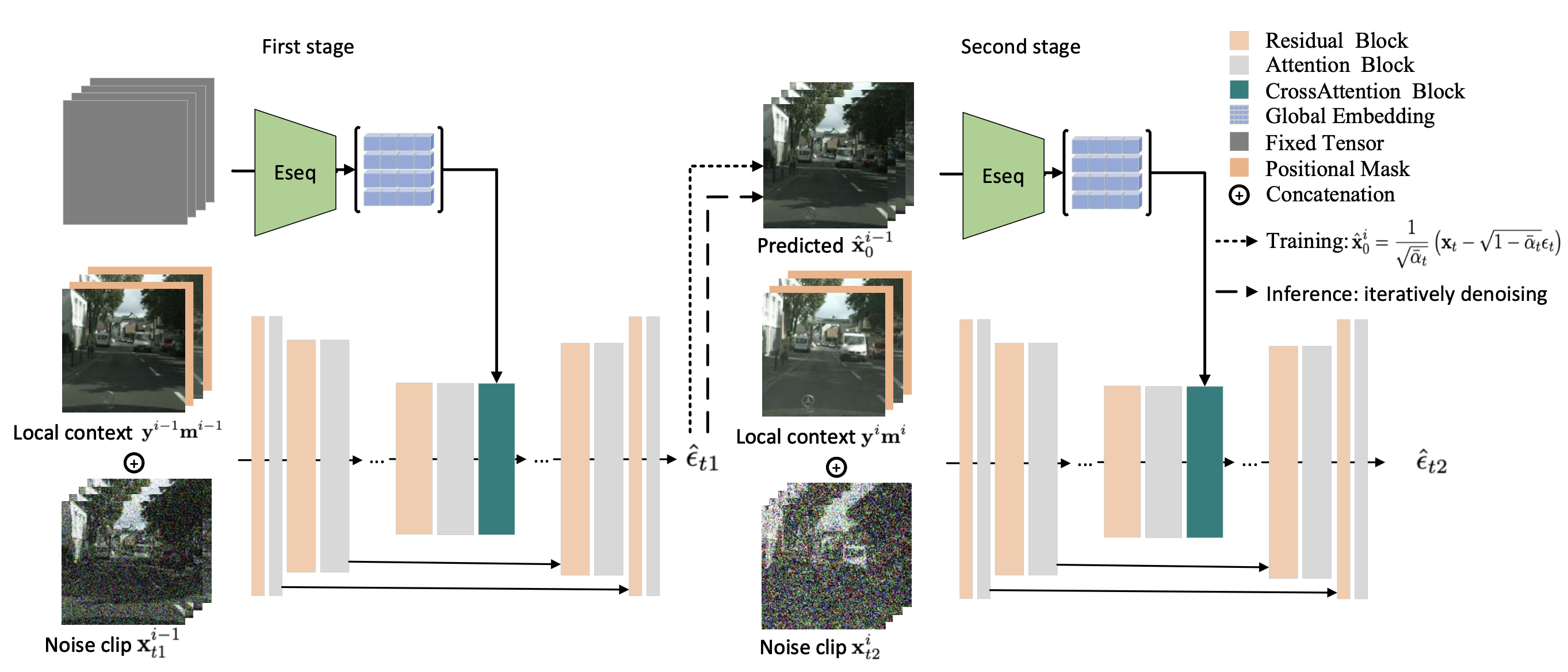}
\caption{Left: In the first stage, the global sequence encoder extracts global context embeddings $\mathbf{z}^{i-1}$ from a fixed tensor. Conditional frames are selected from the ground truth, and after concatenation with positional masks, local context $\mathbf{y^{i-1}}\mathbf{m^{i-1}}$is obtained. 
Right: In the second stage, the sequence encoder extracts global context embeddings from the entire predicted $\mathbf{\hat{x}}^{i-1}_0$, and the conditional frames are selected from the predicted $\mathbf{\hat{x}}^{i-1}_0$. In the test phase, our model performs iterative denoising on Gaussian noise to obtain the prediction of the current fragment $\mathbf{x}_0$,  after which it moves on to the following fragment. In the training phase, our model processes the noisy current fragment $\mathbf{x}_t$ and predicts the noise $\epsilon_t$ to obtain $\mathbf{x}_0$ directly by Eq.\ref{x_0}.
}
\label{architecture}
\end{figure*}

\paragraph{Generative Models for Video Synthesis.} 
Video synthesis is the task of predicting a sequence of visually consistent frames under the condition of text description, prior frames, or Gaussian noises. To remedy this problem, early methods~\cite{vRNN1,vRNN2,vRNN3} build the image-autoregressive model on Recurrent Neural Networks (RNNs) to model the temporal correspondence implicitly. For example, a stochastic video prediction model is proposed in~\cite{vRNN1}, where the temporal memories embedded in the latent value are used to guide the generation of future frames via a recurrent architecture. 
Later, Franceschi~\emph{et al.}~\shortcite{vRNN4} propose a stochastic dynamic model by building state-space models in a residual framework. Akan~\emph{et al.}~\shortcite{vRNN5} propose to use LSTM to model the temporal memory in a latent space for frame prediction and optical flow estimation, respectively. 
Although impressive breakthrough has been made by the aforementioned methods, the recurrent architectures they used usually bring more computational burdens and show limited capacity for modeling the long-term embedding. To alleviate this issue, some recent methods~\cite{Transformer1,Transformer2,Transformer3,weissenborn2019scaling} use Transformer to capture the global space-time correspondence for more accurate video prediction. To further improve the details of the generated videos, some attempts~\cite{GAN1,GAN7} 
have been made by building video prediction models on Generative Adversarial Networks (GANs). For example, \cite{GAN5} propose MoCoGAN to learn to disentangle motion from content in an unsupervised manner. \cite{GAN6} propose TS-GAN to model spatiotemporal consistency among adjacent frames by building a temporal shift module on a generative network. More recently, DIGAN~\cite{GAN8} builds an implicit neural representation based on GAN to improve the motion dynamics by manipulating the space and time coordinates.

\paragraph{Diffusion Probabilistic Models.} Recently, Diffusion Probabilistic Models (DPMs)~\cite{ho2020denoising} have received increasing attention due to their impressive ability in image generation. It has broken the long-term domination of GANs and become the new state-of-the-art protocol in many computer vision tasks, such as image/video/3D synthesis~\cite{dhariwal2021diffusion,saharia2022photorealistic,ramesh2022hierarchical,nichol2021glide}, image inpainting~\cite{lugmayr2022repaint} and super-resolution \cite{rombach2022high}. 
Based on the predefined step-wise noise in the forward process,  DPMs \cite{ho2020denoising,song2020denoising,nichol2021improved,song2020score} leverage a parametric U-Net to denoise the input Gaussian distribution iteratively in the reverse process. 
For unconditional video generation, \cite{ho2022video} extends the original UNet to 3D format \cite{cciccek20163d} to process video data. 
For conditional video prediction tasks, a naive approach is to use the unconditional video generation model directly by means of conditional sampling as RePaint \cite{lugmayr2022repaint}. However, this approach relies on a well-trained unconditional generative model, which is computationally resource-intensive to obtain.
To overcome this problem, \cite{cciccek20163d} and \cite{harvey2022flexible} propose a diffusion-based architecture for video prediction and infilling. For a video with $m$ frames, they randomly select $n$ frames as the conditional frames, which are kept unchanged in the forward and backward processes, and carry out diffusion and denoising on the remaining $m-n$ frames. 
MCVD proposed by \cite{voleti2022mcvd} concatenates all video frames in channel dimension, and represents the video as 
four-dimensional data. This work uses 2D Conv instead of 3D Conv, which greatly reduces the computational burden without compromising the quality of the results, and achieves SOTA results on video prediction and interpolation.



\section{Method}
In this paper, we propose a Local-Global Context guided Video Diffusion model (LGC-VD) to address video synthesis tasks, including video prediction, interpolation, and unconditional generation. Our model is built upon an autoregressive inference framework, where the frames from the previous prediction are used as conditions to guide the generation of the next clip. We construct a local-global context guidance strategy to achieve comprehensive embeddings of the past fragment to boost the consistency of future prediction. Besides, we propose a two-stage training algorithm to alleviate the influence of noisy conditional frames and facilitate the robustness of the diffusion model for more stable prediction. Below, we first review the diffusion probabilistic models.  

\subsection{Background}
Given a sample from data distribution $\mathbf{x}_0 \sim q(\mathbf{x})$, the forward process of the diffusion model destroys the structure in data by adding Gaussian noise to $\mathbf{x}_0$ iteratively. The Gaussian noise is organized according to a variance schedule $\beta_1,...,\beta_T$, resulting in a sequence of noisy samples $\mathbf{x}_1,...\mathbf{x}_T$, where $T$ is diffusion step. When $T \to \infty$, $\mathbf{x}_T$ is equivalent to an isotropic Gaussian distribution. This forward process can be defined as 
\begin{align}
q\left(\mathbf{x}_{1: T}\right) &:=\prod_{t=1}^T q\left(\mathbf{x}_t \mid \mathbf{x}_{t-1}\right)   \\
q\left(\mathbf{x}_t \mid \mathbf{x}_{t-1}\right) &:=\mathcal{N}\left(\mathbf{x}_t ; \sqrt{1-\beta_t} \mathbf{x}_{t-1}, \beta_t \mathbf{I}\right) 
\end{align}
With the transition kernel above, we can sample $\mathbf{x}_t$ at any time step $t$:
\begin{equation}
\begin{aligned}
q_t\left(\mathbf{x}_t \mid \mathbf{x}_0\right)=\mathcal{N}\left(\mathbf{x}_t ; \sqrt{\bar{\alpha}_t} \mathbf{x}_0,\left(1-\bar{\alpha}_t\right) \mathbf{I}\right) \label{xt} 
\end{aligned}
\end{equation}
where $\bar{\alpha}_t=\prod_{s=1}^t\left(1-\beta_s\right)$. In the training phase, the reverse process tries to trace back from the isotropic Gaussian noise $\mathbf{x}_T \sim \mathcal{N}\left(\mathbf{x}_T ; \mathbf{0}, \mathbf{I}\right)$ to the initial sample $\mathbf{x}_0$. Since the exact reverse distribution $q\left(\mathbf{x}_{t-1} \mid \mathbf{x}_t\right)$ cannot be obtained, we use a Markov chain with learned Gaussian transitions to replace it:
\begin{align}
p_\theta\left(\mathbf{x}_{0: T}\right) &:=\prod_{t=1}^T p_\theta\left(\mathbf{x}_{t-1} \mid \mathbf{x}_t\right) \\
p_\theta\left(\mathbf{x}_{t-1} \mid \mathbf{x}_t\right) &:=\mathcal{N}\left(\mathbf{x}_{t-1} ; \boldsymbol{\mu}_\theta\left(\mathbf{x}_t, t\right), \mathbf{\Sigma}_\theta\left(\mathbf{x}_t, t\right)\right) 
\end{align}

In practice, we use KL divergence to get $p_\theta\left(\mathbf{x}_{t-1} \mid \mathbf{x}_t\right)$ to estimate the corresponding forward process posteriors :
\begin{equation}
q\left(\mathbf{x}_{t-1} \mid \mathbf{x}_t, \mathbf{x}_0\right)=\mathcal{N}\left(\mathbf{x}_{t-1} ; \tilde{\boldsymbol{\mu}}_t\left(\mathbf{x}_t, \mathbf{x}_0\right), \tilde{\beta}_t \mathbf{I}\right) \label{reverse} 
\end{equation}
where $ \tilde{\boldsymbol{\mu}}_t\left(\mathbf{x}_t, \mathbf{x}_0\right):=\frac{\sqrt{\bar{\alpha}_{t-1}} \beta_t}{1-\bar{\alpha}_t} \mathbf{x}_0+\frac{\sqrt{\alpha_t}\left(1-\bar{\alpha}_{t-1}\right)}{1-\bar{\alpha}_t} \mathbf{x}_t$, $\alpha_t=1-\beta_t$ and $\tilde{\beta}_t:=\frac{1-\bar{\alpha}_{t-1}}{1-\bar{\alpha}_t} \beta_t$. Since $\mathbf{x}_0$ given $\mathbf{x}_T$ is unknown in the sampling phase, We use a time-conditional neural network parameterized by $\theta$ to estimate the noise $\epsilon_t$, then we can get 
\begin{equation}
\hat{\mathbf{x}}_0=\frac{1}{\sqrt{\bar{\alpha}_t}}\left(\mathbf{x}_t-\sqrt{1-\bar{\alpha}_t} \epsilon_t\right) 
\label{x_0}
\end{equation}
by Eq.\ref{xt}. The loss can be formulated as  
\begin{equation}
L(\theta)=\boldsymbol{E}_{\mathbf{x}_0, \boldsymbol{\epsilon}}\left[\left\|\boldsymbol{\epsilon}-\boldsymbol{\epsilon}_\theta\left(\sqrt{\bar{\alpha}_t} \mathbf{x}_0+\sqrt{1-\bar{\alpha}_t} \boldsymbol{\epsilon}, t\right)\right\|^2\right] 
\label{eq:dpmloss}
\end{equation}
Eventually, we can reverse the diffusion process by replacing $\mathbf{x}_0$ in Eq.\ref{reverse} with $\hat{\mathbf{x}}_0$ to reconstruct the structure in data from Gaussian noise.

$$
\mathbf{y}^{i-1}\mathbf{m}^{i-1} , \mathbf{x}^{i-1}_{t1},
\hat{\mathbf{x}}^{i-1}_0, \mathbf{y}^{i}\mathbf{m}^{i}, \mathbf{x}^{i}_{t2},
\hat{\mathbf{\epsilon}}_{t1},\hat{\mathbf{\epsilon}}_{t2}
$$

\subsection{LGC-VD for Video Synthesis Tasks}

In contrast to \cite{voleti2022mcvd}, our model extracts both local and global context from the past fragment, to boost the consistency of future prediction. For video prediction task, given $p$ start frames, our model aims to produce a fragment with $k$ frames each time and finally propose a $(n=k\times m+p)$-length video, where $m$ is the number of fragments. 

As shown in Figure~\ref{architecture}, to predict the fragment $\mathbf{x}^{i}$ at time $i\in \{0,...,m-1\}$, we use the local context from last fragment $\mathbf{x}^{i-1}$ as local condition, which is represented as $\mathbf{y}^{i}$. 
Besides, we employ a sequential encoder to extract the global feature from the last fragment as a global condition, which is represented as $\mathbf{z}^i$.
Since no past fragment is given at $i=0$, the sequential encoder takes only the input as a fixed tensor $\mathbf{U}$ that shares the same shape as $\mathbf{x}^{i}$. 
The local condition and the global embedding are incorporated in the UNet to guide the generation of the future fragment. 

%
We introduce a positional mask on local conditional frames for the more flexible switch to different tasks. Specifically, each selected conditional frame is concatenated with a positional mask $\mathbf{m} \in \mathbf{R}^{1\times H\times W}$, whose pixels are filled with a fixed value of $\frac{j+1}{k+1}, j \in [-1, k]$. Then, our model can take any frame from the fragment as a condition and flexibly conduct future prediction or infilling. Besides, for unconditional generation, we set $j=-1$ to indicate a fixed tensor $\mathbf{U}$ as initial conditions. Finally, the Eq.~\ref{eq:dpmloss} can be rewritten as    
\begin{equation}
L(\theta)=\boldsymbol{E}_{\mathbf{x}^i, \boldsymbol{\epsilon}}\left[\left\|\boldsymbol{\epsilon}-\boldsymbol{\epsilon}_\theta\left(\sqrt{\bar{\alpha}_t} \mathbf{x}^i+\sqrt{1-\bar{\alpha}_t} \boldsymbol{\epsilon}, t, \mathbf{y}^i\mathbf{m}^i, \mathbf{z}^i \right)\right\|^2\right] 
\end{equation}
Where $ \mathbf{y}^i\mathbf{m}^i$ is the concatenation of $\mathbf{y}^i$ and $\mathbf{m}^i$. In practice, the local condition $\mathbf{y}\mathbf{m}$ is incorporated into the network via channel-wise concatenation, whereas the global condition $\mathbf{z}$ is mapped to the intermediate layers of the UNet via a cross attention layer: 
\begin{align}
\operatorname{Attention}(Q, K, V)=\operatorname{softmax}\left(\frac{Q K^T}{\sqrt{d}}\right) \cdot V \\
 Q=W_Q^{(i)} \cdot \mathbf{f}^{i}, K=W_K^{(i)} \cdot \mathbf{z}^i, V=W_V^{(i)} \cdot \mathbf{z}^i
\end{align}
where $\mathbf{z}^i$ is the feature of $\mathbf{x}^{i-1}$ by sequential encoder. 
$\mathbf{f}^{i}$ is the feature of current fragment by UNet encoder and $W_Q^{(i)},W_K^{(i)},W_V^{(i)}$ are learnable projection matrices.

\begin{figure}[t]
\centering
\includegraphics[width=\linewidth]{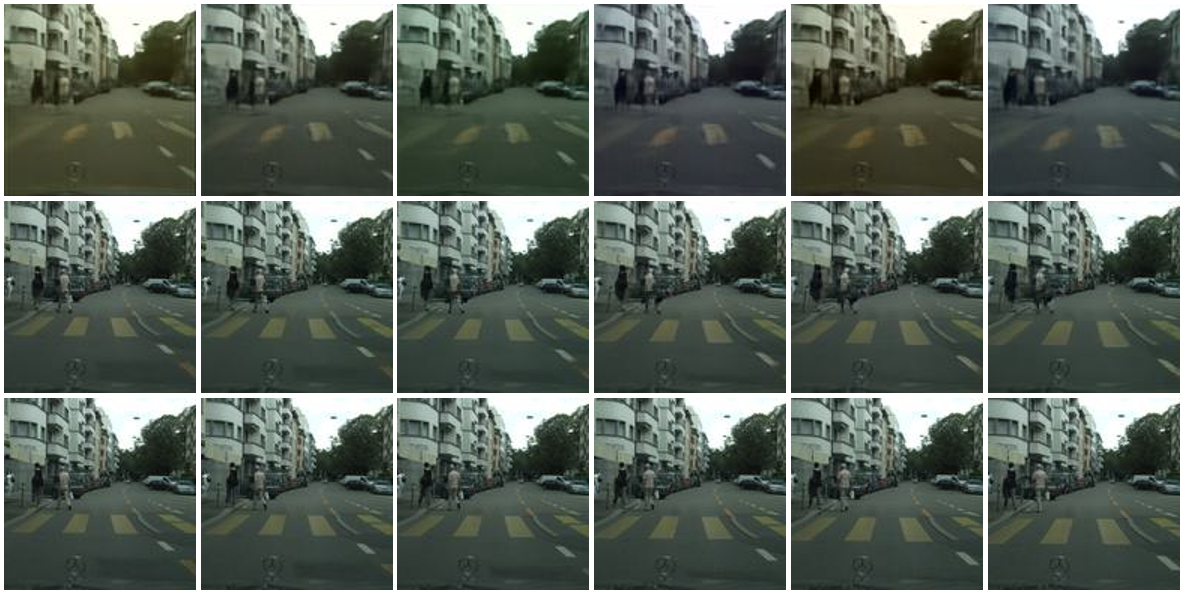}
\caption{Top row shows the output of the first stage, which is referring to the start frames, adding noise to current fragment $\mathbf{x}_0$ using Eq.\ref{xt} to obtain noisy fragment $\mathbf{x}_{592}$, then predicting $\mathbf{x}_0$ directly using Eq.\ref{x_0}.
The middle row shows the prediction of $\mathbf{x}_0$ obtained by iterative denoising from Gaussian noise using the start frames as a reference. The third row shows the ground truth.
The six columns are frames 2-7. 
}
\label{mid output}
\end{figure}

\begin{table*}[thp]
    \centering
        \begin{tabular}{l|rrr|ccc}
        \toprule Cityscapes & $p$ & $k$ & $n$ & FVD $\downarrow$ & LPIPS $\downarrow$ & SSIM $\uparrow$ \\
        \midrule 
        \textbf{LGC-VD} (Ours)    & $1$ & $7$ &$28$ & $182.8$ &$0.084\pm0.03$ & $0.693 \pm 0.08$ \\
        \midrule SVG-LP \cite{denton2018stochastic} & $2$ & $10$ &$28$  & $1300.26$ & $0.549 \pm 0.06$ & $0.574 \pm 0.08$ \\
        vRNN 1L \cite{castrejon2019improved} & $2$ & $10$ &$28$ & $682.08$ & $0.304 \pm 0.10$ & $0.609 \pm 0.11$ \\
        Hier-vRNN \cite{castrejon2019improved} & $2$ & $10$ &$28$ & $567.51$ & $0.264 \pm 0.07$ & $0.628 \pm 0.10$ \\
        GHVAE \cite{wu2021greedy} & $2$ & $10$ &$28$ & $418.00$ & $0.193 \pm 0.014$ & $\mathbf{0 . 7 4 0} \pm 0.04$ \\
        MCVD spatin \cite{voleti2022mcvd}  & $2$ & $5$ &$28$  & $184.81$ & $0.121 \pm 0.05$ & $0.720 \pm 0.11$ \\
        MCVD concat \cite{voleti2022mcvd} & $2$ & $5$ &$28$ & $141.31$ & $0.112 \pm 0.05$ & $0.690 \pm 0.12$ \\
        \midrule
        \textbf{LGC-VD} (Ours)    & $2$ & $6$ &$28$ & $\mathbf{124.62}$ & $\mathbf{0.069} \pm 0.03$ & $0.732 \pm 0.09$ \\
        \bottomrule
        \end{tabular}
    \caption{Video prediction on Cityscapes. Predicting $k$ frames using the first $p$ frames as a condition, then recursively predicting $n$ frames. We illustrate video prediction results under various lengths of conditions by $p=1$ and $p=2$.}
    \label{cityscape prediction}
\end{table*}

\subsection{Two-stage Training}
In previous autoregressive frameworks~\cite{voleti2022mcvd,ho2022video,hoppe2022diffusion}, the ground truths of previous frames are directly used as conditions during training. However, undesirable errors in content or motion would accumulate and affect long-term predictions. To alleviate this issue, we propose a two-stage training strategy.

In the first stage, the conditional frames $\textbf{y}^0$ are randomly selected from ground truth $\textbf{x}^0$, and the sequential encoder takes the input as the constant tensor $\textbf{U}$. The model then outputs the first predicted video sequence $\hat{\textbf{x}}^0$. In the second stage, the conditional frames $\textbf{y}^1$ are the last few frames from the previous prediction $\hat{\textbf{x}}^0$, the conditional feature $\textbf{z}^0$ is the result of encoding $\hat{\textbf{x}}^0$ by the sequence encoder. With this setting, the prediction errors in the training phase are included and are treated as a form of data augmentation to improve the network's ability for long video prediction. Therefore, the first stage of training endows our model with the flexibility to make future predictions or infillings, and the second stage enhances the model's ability to predict long videos.

Since iteratively denoising in the training phase will greatly increase the computational burden, we use Eq.\ref{x_0} to directly obtain the prediction of the current fragment $\hat{\mathbf{x}}_0$ from the noisy fragment $\mathbf{x}_t$. A natural question is whether using the $\hat{\mathbf{x}}_0$ obtained through Eq.\ref{x_0} instead of the $\hat{\mathbf{x}}_0$ obtained by iterative denoising can help the model learn to combat prediction error. To evaluate this, we randomly select a training sample and display the results obtained by Eq.\ref{x_0} (top row) and the results obtained by iterative denoising (middle row) in Figure \ref{mid output}. As can be seen, the results of both predictions show blurry effects in uncertainty areas (\emph{e.g,} the occluded region of the background or the regions with object interaction). Therefore, this substitution can be considered a legitimate data enhancement.

\section{Experiments}
We present the experimental results of \textbf{video prediction} on \textbf{Cityscapes} \cite{cordts2016cityscapes}, and the experimental results of \textbf{video prediction}, \textbf{video generation}, \textbf{video interpolation} on \textbf{BAIR} \cite{ebert2017self}.

\subsection{Datasets and Metrics}
\paragraph{Dataset.} Cityscapes~\cite{cordts2016cityscapes} is a large-scale dataset that contains a diverse set of stereo video sequences recorded in street scenes from 50 different cities. We use the same data preprocessing method as \cite{yang2022diffusion,voleti2022mcvd}, from the official website, we obtain the \textit{leftImg8bit\_sequence\_trainvaltest}. This package includes a training set of 2975 videos, a validation set of 500 videos, and a test set of 1525 videos, each with 30 frames. All videos are center cropped and downsampled to 128x128. 
BAIR Robot Pushing~\cite{ebert2017self} is a common benchmark in the video literature, which consists of roughly 44000 movies of robot pushing motions at 64x64 spatial resolution.

\paragraph{Metrics.}
In order to compare our LGC-VD with prior works, we measure our experimental results in PSNR, LPIPS, SSIM, and FVD \cite{unterthiner2018towards}. 
FVD compares statistics in the latent space of an Inflated 3D ConvNet (I3D) trained on Kinetics-400, which measures time coherence and visual quality. 

\subsection{Implementation Details}
We use a U-Net that is composed of encoding and decoding for upsample and downsample, respectively. Each part contains two residual blocks and two attention blocks. The middle layer of UNet is composed of one residual block followed by an attention block. The sequential encoder uses the front half and middle layer of diffusion UNet, with a total of three residual blocks and three attention modules. Since our sequential encoder only needs to be called once when iteratively denoising, the added time in the test phase is negligible. Besides, in our experiments, we found that when $\epsilon-prediciton$ is coupled with our model's spatial-temporal attention block, particularly when trained at resolutions of $128 \times 128$ and higher, the predicted video has brightness variations and occasionally color inconsistencies between frames. We use v-prediction \cite{salimans2022progressive} to overcome this problem. 

All of our models are trained with Adam on 4 NVIDIA Tesla V100s with a learning rate of 1e-4 and a batch size of 32 for Cityscapes and 192 for BAIR. We use the cosine noise schedule in the training phase and set the diffusion step $T$ to 1000. For both datasets, we set the total video length $L$ to 14, the video length $N$ for each stage to 8, and the number of conditional frames $K$ to 2. At testing, we sample 100 steps using DDPM.

\subsection{Evaluation on Video Prediction}

Table \ref{cityscape prediction} lists all the metric scores of our model and other competitors on Cityscapes. For video prediction, our model significantly outperforms MCVD~\cite{voleti2022mcvd} in FVD, LPIPS, and SSIM. Despite the fact that our SSIM is slightly lower than GHVAE~\cite{wu2021greedy}, it significantly outperforms GHVAE in FVD and LPIPS.

Table \ref{bair prediction} lists all the metric scores of our model and other methods on BAIR. To compare our model with previous work on video prediction, we evaluated our model under two settings of start frames. With the first two frames serving as references, the two evaluation methods predict 14 frames and 28 frames, respectively. As can be seen, our model is slightly lower than MCVD in terms of FVD, but much better than MCVD in terms of PSNR and SSIM. Besides, Table \ref{bair prediction} also shows that our model's result under just one start frame ($p=1$), which outperforms other methods by large margin. 

\begin{figure*}[!h]
\centering
\includegraphics[width=\linewidth]{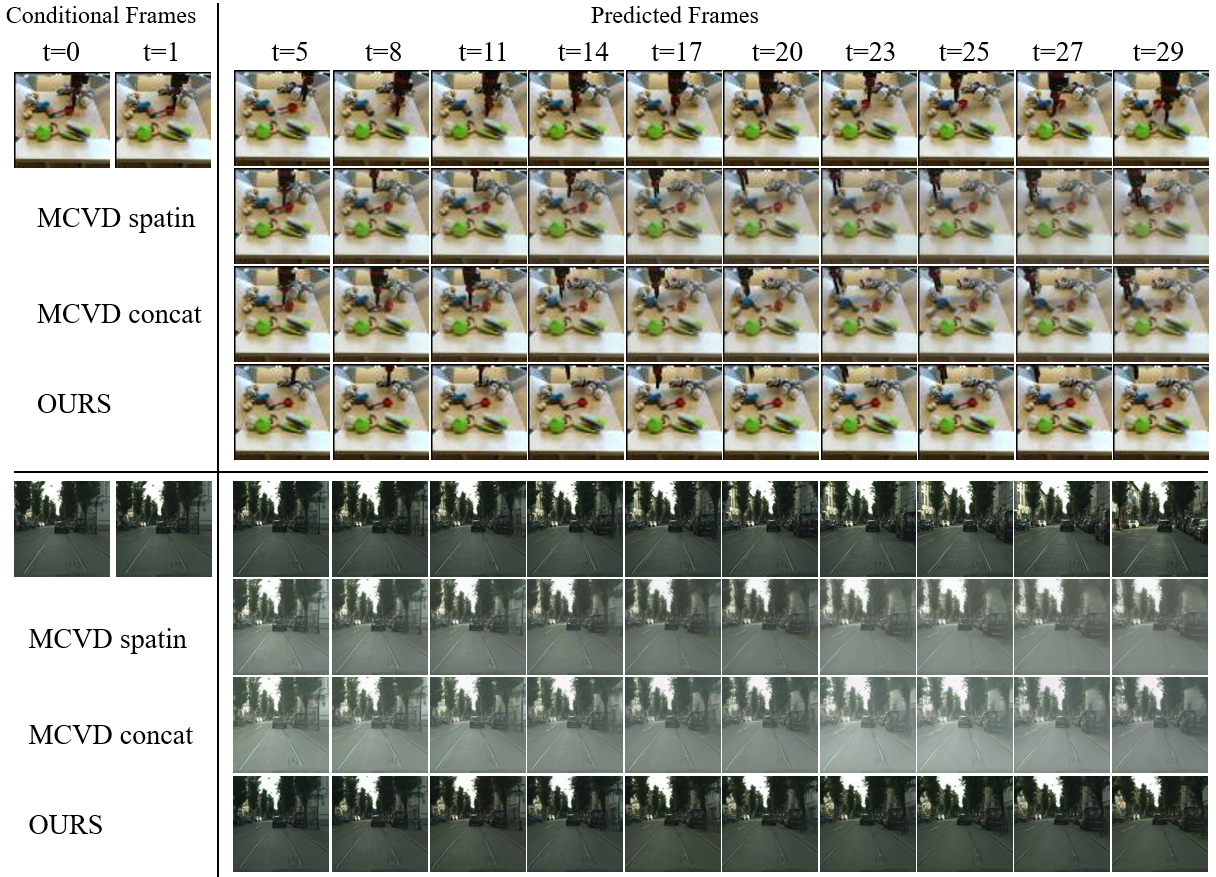}
\caption{ Qualitative results of video prediction on Cityscapes and BAIR. From top to bottom: Ground Truth (Top Row), MCVD spatin (Second Row), MCVD concat (Third Row), Our Method (Bottom Row). On BAIR, we observe that the MCVD prediction becomes blurry over time while our LGC-VD predictions remain crisp. On Cityscapes, the prediction of MCVD shows significant brightness biases, while our LGC-VD performs better in both motion and content consistency.}
\label{figure:visualexamples}
\end{figure*}

The perceptual performance is also verified visually in Figure \ref{figure:visualexamples}. As we can see, our LGC-VD is effective at producing high-quality videos with clear details, while MCVD tends to produce blurry results on the regions with potentially fast motion. Besides, our LGC-VD alleviates the issue of inconsistent brightness in MCVD.

\subsection{Evaluation on Video Interpolation}
The experimental results of our model's video interpolation on BAIR are shown in Table \ref{bair interpolation}. Compared with MCVD, our model requires fewer reference images, predicts more intermediate frames, and outperforms MCVD in terms of SSIM and PSNR. Even when compared to architectures designed for video interpolation \cite{xu2020stochastic,niklaus2017video}, our model obtains SOTA results for both metrics, 26.732 and 0.952, respectively.
\begin{table}[!h]
    \centering
        \begin{tabular}{l|r|r}
        \toprule BAIR & $p+f$ / $k$  / $n$ & PSNR$\uparrow$  / SSIM $\uparrow$\\
        \midrule SVG-LP  & $18$ / $7$ / $100$  & $18.648$ / $0.846$ \\
       SepConv  & $18$ / $7$ / $100$ & $21.615$ / $0.877$ \\
       SDVI full  & $18$ / $7$ / $100$  & $21.432$ / $0.880$ \\
       SDVI  & $18$ / $7$ / $100$  & $19.694$ / $0.852$ \\
       MCVD  & $4$ / $5$ / $100$  & $25.162$ / $0.932$ \\
       \midrule
        \textbf{LGC-VD} (ours)  & $2$ / $6$ / $100$ & $\mathbf{26.732} / \mathbf{0.952}$\\
        \bottomrule 
        \end{tabular}
    \caption{Video interpolation on BAIR. Given $p$ past with $f$ future frames, interpolating $k$ frames. We report average of the best metrics out of $n$ trajectories per test sample. }
    \label{bair interpolation}
\end{table}

\subsection{Evaluation on Video Generation}
\begin{table*}[!h]
    \centering
        \begin{tabular}{l|rrr|ccc}
        \toprule BAIR  & $p$ & $k$ & $n$ & FVD $\downarrow$ & PSNR $\uparrow$ & SSIM $\uparrow$ \\
        \midrule LVT \cite{Transformer3}  & 1 & 15 & 15 & $125.8$ & $-$ & $-$ \\
        DVD-GAN-FP \cite{clark2019adversarial} & 1 & 15 & 15 & $109.8$ & $-$ & $-$ \\
        TrIVD-GAN-FP \cite{luc2020transformation} & 1 & 15 & 15 & $103.3$ & $-$ & $-$ \\
        VideoGPT \cite{Transformer1} & 1 & 15 & 15 & $103.3$ & $-$ & $-$ \\
        CCVS  \cite{Transformer2} & 1 & 15 & 15 & $99.0$ & $-$ & $-$ \\
        Video Transformer \cite{weissenborn2019scaling}  & 1 & 15 & 15 & $96$ & $-$ & $-$ \\
        FitVid \cite{babaeizadeh2021fitvid} & 1 & 15 & 15 & $93.6$ & $-$ & $-$ \\
        MCVD spatin past-mask \cite{voleti2022mcvd} & 1 & $5$ & 15 & $96.5$ & $18.8$ & $0.828$ \\
        MCVD concat past-mask \cite{voleti2022mcvd} & 1 & $5$ & 15 & $95.6$ & $18.8$ & $0.832$ \\
        MCVD concat past-future-mask \cite{voleti2022mcvd} & 1 & $5$ & 15 & $89.5$ & $16.9$ & $0.780$ \\
        \midrule 
        \textbf{LGC-VD} (Ours)  & 1 & $7$ & 15 & $\mathbf{80.9}$ & $\mathbf{21.8}$ & $\mathbf{0.891}$ \\
        \midrule SAVP \cite{lee2018stochastic}  & 2 & 14 & 14 & $116.4$ & $-$ & $-$ \\
        MCVD spatin past-mask \cite{voleti2022mcvd} & 2 & $5$ & 14 & $90.5$ & $ 19.2$ & $0.837$ \\
        MCVD concat past-future-mask \cite{voleti2022mcvd} & 2 & $5$ & 14 & $89.6$ & $17.1$ & $0.787$ \\
        MCVD concat past-mask \cite{voleti2022mcvd} & 2 & $5$ & 14 & $87.9$ & $19.1$ & $0.838$ \\
        \midrule 
        \textbf{LGC-VD} (Ours)   & 2 & $6$ & 14 & $\mathbf{76.5}$ & $\mathbf{21.9}$ & $\mathbf{0.892}$ \\
        \midrule SVG-LP \cite{denton2018stochastic} & 2 & 10 & 28 & $256.6$ & $-$ & $0.816$ \\
        SLAMP \cite{vRNN5} & 2 & 10 & 28 & $245.0$ & $19.7$ & $0.818$ \\
        SAVP \cite{lee2018stochastic} & 2 & 10 & 28 & $143.4$ & $-$ & $0.795$ \\
        Hier-vRNN \cite{castrejon2019improved} & 2 & 10 & 28 & $143.4$ & $-$ & $0.822$ \\
        MCVD spatin past-mask \cite{voleti2022mcvd} & 2 & $5$ & 28 & $127.9$ & $17.7$ & $0.789$ \\
        MCVD concat past-mask \cite{voleti2022mcvd} & 2 & $5$ & 28 & $119.0$ & $17.7$ & $0.797$ \\
        MCVD concat past-future-mask \cite{voleti2022mcvd} & 2 & $5$ & 28 & $\mathbf{118.4}$ & $16.2$ & $0.745$ \\
        \midrule 
        \textbf{LGC-VD} (Ours)   &$2$ &$6$ &$28$  & $120.1$ & $\mathbf{20.39}$ & $\mathbf{0.863}$ \\
        \bottomrule
        \end{tabular}
    \caption{Video prediction on BAIR. Predicting $k$ frames using the first $p$ frames as a condition, then recursively predicting $n$ frames in total.}
    \label{bair prediction}
\end{table*}

We also present the results of unconditional video generation on BAIR. In this setting, no conditional frames are given and the models need to synthesize the videos from only Gaussian noise. As shown in Table \ref{bair generation}, our model first generates 8-frame video sequences in an unconditional manner, then generates 6-frame video sequences each time using our autoregressive inference framework, and finally creates video sequences with 30 frames. Our model yields a FVD of $250.71$, which is significantly improved compared to the $348.2$ of MCVD.
\begin{table}[!h]
    \centering
        \begin{tabular}{l|cc|c}
        \toprule  BAIR & $k$ & $n$ & FVD $\downarrow$ \\
        \midrule MCVD spatin \cite{voleti2022mcvd} & $5$ & $30$  & $399.8$ \\
        MCVD concat \cite{voleti2022mcvd}  & $5$ & $30$ & $348.2$ \\
        \midrule 
        \textbf{LGC-VD} (ours)  & $6$ & $30$ & $\mathbf{250.7}$    \\
        \bottomrule 
        \end{tabular}
    \caption{Unconditional video generation on BAIR. Generating $k$ frames every time, then recursively predicting $n$ frames.}
    \label{bair generation}
\end{table}
\subsection{Ablation Study}
The proposed LGC-VD is built upon an autoregressive inference framework, where several past frames are taken as a condition to boost the generation of the next video fragment. In LGC-VD, we propose a hierarchical context guidance strategy to incorporate the local context from a few past frames with the global context of historical clips. Besides, considering that the prediction would be affected by the noisy conditions, a two-stage training algorithm is proposed to facilitate the robustness of the model for more stable prediction. 
To demonstrate the efficacy of the two contributions, we conduct ablation studies on Cityscapes. As shown in Table \ref{ablation result}, we implement a baseline by removing the sequential encoder and the two-stage training strategies. Specifically, the UNet takes only the past $p$ frames as a condition to model the local guidance, and the second training is skipped. From the comparison among ``$+$two-stage training'', ``$+$sequential encoder'' and ``baseline'', we can observe that the proposed two-stage training strategy and local-global context guidance can significantly improve the stability of video prediction. By combining the sequential encoder and the local-global context guidance, our model achieves further improvement, especially on FVD, which demonstrates that the global memory from the sequential encoder can collaborate with the original local context to achieve more consistent results. 
\begin{table}[!h]
    \centering
        \begin{tabular}{l|r|rll}
        \hline Cityscapes & $p$ /$k$ / $n$ & FVD  / LPIPS  / SSIM  \\
        \hline Baseline & $2$/$6$/$28$ &  $276.27$ / $0.111 $ / $0.708 $ \\
        $+$ two-stage training & $2$/$6$/$28$ &  $141.91$ / $0.071 $ / $0.742 $ \\
        $+$ sequential encoder & $2$/$6$/$28$ &  $153.08$ / $\mathbf{0.067} $ / $\mathbf{0.750} $ \\
        full    & $2$/$6$/$28$ &  $\mathbf{124.62}$ / $0.069$ / $0.732 $ \\
        \hline
        \end{tabular}
    \caption{Ablation study on Cityscape. Note that we conduct experiments on video prediction, where 2 initial frames ($p=2$) are conditions to predict 6 frames (k=6) and propose a 28-frame video ($n=28$).}
    \label{ablation result}
\end{table}

\section{Conclusion and Discussion}
In this paper, we propose local-global context guidance to comprehensively construct the multi-perception embedding from the previous fragment, for high-quality video synthesis. We also propose a two-stage training strategy to alleviate the effect of noisy conditions and boost the model to produce more stable predictions.
Our experiments demonstrate that the proposed method achieves state-of-the-art performance on video prediction, as well as favorable performance on interpolation and unconditional video generation. 
To summarize, our method makes a further improvement in the condition formulation and the training stability for memory-friendly video diffusion methods. 

\section*{Acknowledgements}
This study was co-supported by the National Key R\&D Program of China under Grant 2021YFA0715202, the National Natural Science Foundation of China (Nos. 62293544, 62022092), as well as partially supported by the National Postdoctoral Program of China for Innovative talent (BX2021051) and the National Nature Science Foundation of China (62206039).


\bibliographystyle{named}
\bibliography{ijcai23}

\end{document}